\documentclass{article} % For LaTeX2e
\usepackage{iclr2019_conference,times}
\usepackage{amsmath,amsfonts,bm}

\def\eqref#1{equation~\ref{#1}}
\def\1{\bm{1}}

\def\vd{{\bm{d}}}

\def\vh{{\bm{h}}}

\def\vx{{\bm{x}}}
\def\vy{{\bm{y}}}

\def\mA{{\bm{A}}}
\def\mB{{\bm{B}}}

\def\mD{{\bm{D}}}

\def\mU{{\bm{U}}}
\def\mV{{\bm{V}}}

\DeclareMathAlphabet{\mathsfit}{\encodingdefault}{\sfdefault}{m}{sl}
\SetMathAlphabet{\mathsfit}{bold}{\encodingdefault}{\sfdefault}{bx}{n}

\def\gE{{\mathcal{E}}}

\def\gG{{\mathcal{G}}}

\def\gK{{\mathcal{K}}}

\def\gV{{\mathcal{V}}}

\def\sS{{\mathbb{S}}}

\def\sU{{\mathbb{U}}}

\def\sX{{\mathbb{X}}}
\def\sY{{\mathbb{Y}}}

\usepackage{url}
\usepackage[pdftex]{graphicx}
\usepackage{textcase}
\usepackage{hyperref}
\usepackage{float}
\usepackage{natbib}

\title{GraphTSNE: A Visualization Technique for Graph-Structured Data}

\author{Yao Yang Leow\\
School of Computer Science and Engineering\\
Nanyang Technological University, Singapore\\
\texttt{yleow002@e.ntu.edu.sg} \\
\And
Thomas Laurent \\
Department of Mathematics \\
Loyola Marymount University \\
\texttt{tlaurent@lmu.edu}
\And
Xavier Bresson \\
School of Computer Science and Engineering\\
Nanyang Technological University, Singapore\\
\texttt{xbresson@ntu.edu.sg} \\
}

\newcommand{\bigO}{\mathcal{O}}
\newcommand{\restatableeq}[3]{\label{#3}#2\gdef#1{#2\tag{\ref{#3}}}}

\AtBeginDocument{\let\latexlabel\label}

\iclrfinalcopy % Uncomment for camera-ready version, but NOT for submission.
\begin{document}

\maketitle

\begin{abstract}
We present GraphTSNE, a novel visualization technique for graph-structured data based on t-SNE. The growing interest in graph-structured data increases the importance of gaining human insight into such datasets by means of visualization. Among the most popular visualization techniques, classical t-SNE is not suitable on such datasets because it has no mechanism to make use of information from the graph structure. On the other hand, visualization techniques which operate on graphs, such as Laplacian Eigenmaps and tsNET, have no mechanism to make use of information from node features. Our proposed method GraphTSNE produces visualizations which account for \emph{both} graph structure and node features. It is based on scalable and unsupervised training of a graph convolutional network on a modified t-SNE loss. By assembling a suite of evaluation metrics, we demonstrate that our method produces desirable visualizations on three benchmark datasets.\footnote{Our implementation is available at: \url{https://github.com/leowyy/GraphTSNE}}
\end{abstract}

%%%%%%%%%%%%%%%%%%%%
\section{Introduction}
%%%%%%%%%%%%%%%%%%%%
Visualization of high-dimensional data has become common practice following the success of dimensionality reduction techniques such as Laplacian Eigenmaps \citep{Belkin01}, t-SNE \citep{Maaten08}, tsNET \citep{Kruiger17}, and UMAP \citep{McInnes18}. In contrast to the more general problem of dimensionality reduction, visualizations can be particularly useful as a tool to explore and hypothesize about given data. 

\begin{figure}[h]
\begin{center}
\includegraphics[width=1.0\linewidth]{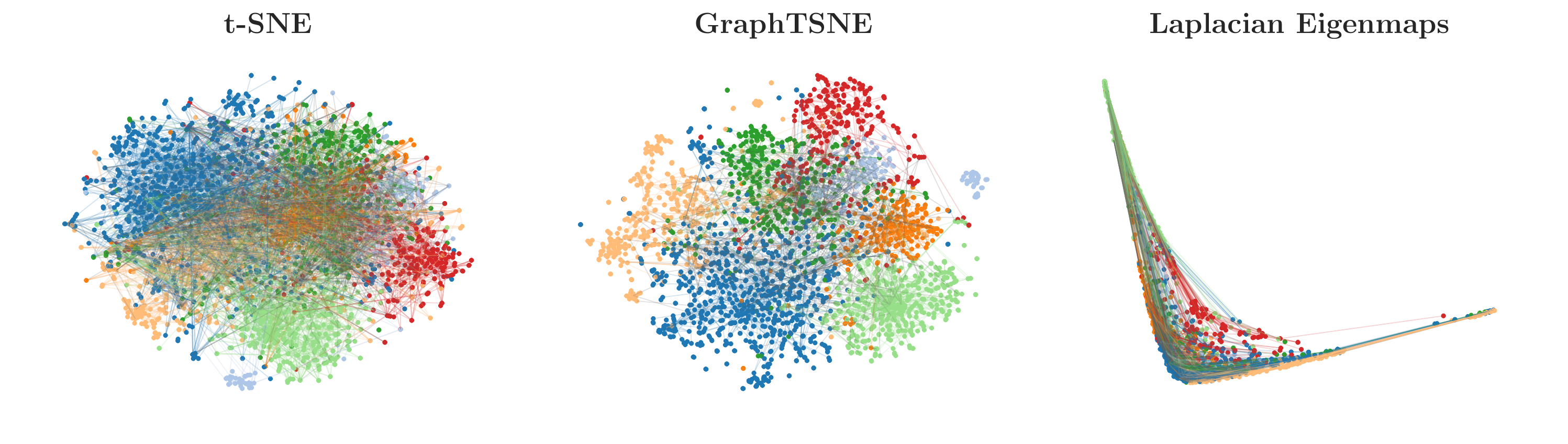}
\end{center}
\caption{Three different visualizations of the \textsc{Cora} citation network. Compared to t-SNE (\textbf{left}) and Laplacian Eigenmaps (\textbf{right}), our proposed method GraphTSNE (\textbf{middle}) is able to produce visualizations which account for \emph{both} graph structure and node features.}
\label{motivation}
\end{figure}

However, to the best of our knowledge, such techniques have not been extended to handle high-dimensional data lying within an explicit graph structure. This limitation results in poor performance on graph-structured datasets as depicted in Figure~\ref{motivation}. Our proposed method GraphTSNE is based on extending the highly successful method of t-SNE to produce visualizations which account for \emph{both} graph structure and node features. In particular, we shall consider datasets with two sources of information: graph connectivity between nodes and node features. Examples of graph-structured datasets include social networks, functional brain networks and gene-regulatory networks. 

Formally, we consider a graph $\gG=(\gV,\gE)$, consisting of a set of nodes $\gV = \{v_i\}^N_{i=1}$, and a set of edges $\gE=\{(v_i, v_j)\} \subseteq \gV \times \gV$ that maps relations between nodes in $\gV$. In addition, each node is associated with a $n$-dimensional feature vector, given by a set of high-dimensional points $\sX = \{\vx_i \in \mathbb{R}^n\}^N_{i=1}$. The goal of visualization is to map the set of nodes $\gV$ to a set of low-dimensional points $\sY= \{\vy_i \in \mathbb{R}^m\}^N_{i=1}$, where $m\ll n$ and typically $m\in\{2,3\}$. 

%%%%%%%%%%%%%%%%%%%%
\section{Related Work}
%%%%%%%%%%%%%%%%%%%%
\textbf{t-SNE.} t-SNE \citep{Maaten08} operates by defining pairwise joint probabilities $p_{ij}$ of picking a point-pair in high-dimensional space and probabilities $q_{ij}$ of picking a point-pair in the low-dimensional map. The probabilities $p_{ij}$ of picking the pair $(\vx_i, \vx_j)$ are parametrized by a normalized Gaussian distribution with mean given by the corresponding entry $\mD_{i,j}$ of a pairwise distance matrix $\mD$. The pairwise distances $\mD_{i,j}$ are computed by a distance measure in high-dimensional space, typically the squared Euclidean distance, i.e. $\mD_{i,j}=|| \vx_i - \vx_j ||^2_2$. Whereas the probabilities $q_{ij}$ of picking the pair $(\vy_i,\vy_j)$ are parametrized by a normalized Student's t-distribution with mean given by the squared Euclidean distance between low-dimensional data points $|| \vy_i - \vy_j ||^2_2$. The objective of t-SNE is to find a low-dimensional data representation $\sY$ that minimizes the mismatch between the two probability distributions $p_{ij}$ and $q_{ij}$, given by the Kullback-Leiber (KL) divergence $C_{\text{KL}} = \sum_i \sum_j p_{ij} \log \frac{p_{ij}}{q_{ij}}$. The t-SNE loss is minimized via gradient descent on $\frac{\partial C_{\text{KL}}} {\partial \vy_i}$.

\textbf{Application of t-SNE on graphs.} In the domain of graph layouts, force-directed graph drawing \citep{Fruchterman91} considers a node placement such that the Euclidean distance between any pair of nodes in the layout is proportional to their pairwise graph-theoretic distances. Theoretical links between dimensionality reduction and graph layouts have been proposed by \citet{Yang14}. Based on this observation, \citet{Kruiger17} have proposed a graph layout algorithm tsNET which operates similar to t-SNE, except the input distance matrix $\mD$ is computed using the pairwise graph-theoretic shortest-path distances between nodes, i.e. $\mD_{i,j}=\delta_\gG(i,j)$. However, similar to other graph layout algorithms \citep{Landesberger11}, tsNET only takes the graph structure $\gG$ as input and therefore, cannot incorporate information provided by node features $\sX$.

%%%%%%%%%%%%%%%%%%%%
\section{Method}
%%%%%%%%%%%%%%%%%%%%
Our method relies on two modifications to a \textit{parametric} version of t-SNE proposed by \citet{Maaten09}. First, we use a graph convolutional network (GCN) (\citealp{Sukhbaatar16, Kipf16, Hamilton17}) as the parametric model for the non-linear mapping between the high-dimensional data space and the low-dimensional embedding space, i.e. $\sY = f_{\text{GCN}}(\gG,\sX)$. In our present work, we use a two-layer residual gated GCN of \citet{Bresson18} with the following layer-wise propagation model:
\begin{equation} 
\vh_i^{l+1} = \text{ReLU} \Big( \mU^l \vh_i^l + \tfrac{1}{|n(i)|}\sum_{j \rightarrow i}\eta_{ij} \odot \mV^l \vh_j^l \Big) + \vh_i^l \text{ ,}
\end{equation} where $\vh_i^l$ denotes the latent representation of node $v_i$ at layer $l$, with $\vh_i^0 = \vx_i$. $\eta_{ij}=\sigma(\mA^l \vh_i^l + \mB^l \vh_j^l)$ denotes edge gates between the node pair $(v_i,v_j)$, with $\sigma(\cdot)$ as the sigmoid function. $|n(i)|$ denotes the indegree of node $v_i$. The learnable parameters of the model are given by $\mA^l, \mB^l, \mU^l, \mV^l$.

Second, we train our GCN using a modified t-SNE loss composed of two sub-losses: a graph clustering loss $C_\gG$ and a feature clustering loss $C_X$. Both sub-losses follow the same formulation of the classical t-SNE loss, but differ in the choice of distance metric used to compute the input pairwise distance matrix $\mD$ between point-pairs. First, the graph clustering loss $C_\gG$ takes as input the pairwise graph-theoretic shortest-path distances, i.e. $\mD_{i,j}^\gG=\delta_\gG(i,j)$. Second, the feature clustering loss $C_X$ takes as input a suitable distance measure between pairs of node features, e.g. the squared Euclidean distance $\mD_{i,j}^X=|| \vx_i - \vx_j ||^2_2$. Lastly, we explore the tradeoff between the two sub-losses by taking a weighted combination: $C_{T} = \alpha C_{\gG} + (1 - \alpha) C_{X}$. The composite loss $C_{T}$ relies on two hyperparameters: the weight of the graph clustering loss $\alpha$ and a perplexity parameter associated with the classical t-SNE loss which measures the effective number of neighbors of a point \citep{wattenberg2016how}. The perplexity parameter is set to a default value of 30 in all experiments. 

%%%%%%%%%%%%%%%%%%%%
\section{Experiments}
%%%%%%%%%%%%%%%%%%%%
We evaluate our method by producing visualizations of three benchmark citation networks - \textsc{Cora}, \textsc{Citeseer} and \textsc{Pubmed} \citep{Sen08} - where nodes are documents, each associated with a high-dimensional word occurrence feature vector, and edges are citation links. Summary statistics for the benchmark datasets are presented in Table~\ref{first_table}. Full details of our experimental setup are provided in Appendix \ref{appendix_a}. 

\begin{table}[h]
\caption{Datasets statistics}
\label{first_table}
\begin{center}
\begin{tabular}{llrrrr}
\bf Dataset        	&\bf Type  	&\bf Nodes  	&\bf Edges 	&\bf Classes 	&\bf Features
\\ \hline 
\textsc{Cora} 		&Citation  		&2,708 		&5,429 		&7 			&1,433\\
\textsc{Citeseer}	&Citation	 	&3,337 		&4,732 		&6 			&3,703\\
\textsc{Pubmed}	&Citation  		&19,717 		&44,328 		&3			&500\\
\end{tabular}
\end{center}
\end{table}

\subsection{Evaluation metrics}
\label{trust}
Following the literature on dimensionality reduction, we adopt metrics based on how well local neighborhoods are preserved in the low-dimensional map. We proceed with these definitions:
\begin{flalign}
\restatableeq{\eqone}{\text{Graph neighborhood:} \quad &\sS_\gG(v_i,r) = \{v_j \in \gV \mid \delta_\gG(i,j) \leq r\} \text{ ;}}{eq1}\\ 
\restatableeq{\eqtwo}{\text{Feature neighborhood:} \quad &\sS_X(v_i,k) = \{v_j \in \gV \mid \vx_j \in k \text{ nearest neighbors of } \vx_i\} \text{ ;}}{eq2}\\
\restatableeq{\eqthree}{\text{Embedding neighborhood:} \quad &\sS_Y(v_i,k) = \{v_j \in \gV \mid \vy_j \in k \text{ nearest neighbors of } \vy_i\} \text{ .}}{eq3}
\end{flalign}

As a graph neighborhood preservation metric, we adopt the graph trustworthiness $T_\gG(r)$, adapted from \citet{Martins15}, which expresses the extent to which graph neighborhoods $\sS_\gG$ are retained in the embedding neighborhoods $\sS_Y$. As a feature neighborhood preservation metric, we adopt the feature trustworthiness $T_X(k)$, proposed by \citet{Venna06}, which expresses the extent to which feature neighborhoods $\sS_X$ are retained in the embedding neighborhoods $\sS_Y$. Complete details of the trustworthiness measures are provided in Appendix \ref{appendix_b}. 

Next, we introduce two distance-based evaluation metrics specific to the goal of visualization. To begin, denote $\gK$ as the $k$-nearest neighbors graph computed by pairwise distances in the \textit{feature} space for a chosen value of $k$. In our visualizations, a point-pair $(\vy_i, \vy_j)$ would ideally be placed close together in the low-dimensional map if they are either connected by an edge in the graph, i.e. $(i,j) \in \gE$, or have similar features, i.e. $(i,j) \in \gK$. Hence, a reasonable objective in visualization is to minimize the graph-based distance $P_G$ and the feature-based distance $P_X$ defined as follows:

\begin{equation*}
\refstepcounter{equation}\latexlabel{firsthalf}
\refstepcounter{equation}\latexlabel{secondhalf}
P_\gG = \frac{1}{|\gE|} \sum_{(i,j)\in \gE} || \vy_i - \vy_j ||^2_2 \text{ ; }\qquad  P_X = \frac{1}{|\gK|}\sum_{(i,j)\in \gK} || \vy_i - \vy_j ||^2_2 \text{ .}
\tag{\ref*{firsthalf}, \ref*{secondhalf}}
\end{equation*}

Lastly, we also report the generalization accuracy of 1-nearest neighbor (1-NN) classifiers which are trained on the set of low-dimensional points $\sY$ obtained from the visualization and the underlying class labels, following the method used by \citet{Maaten09}. 

\subsection{Results}
\begin{figure}[h]
\begin{center}
\includegraphics[width=1.0\linewidth]{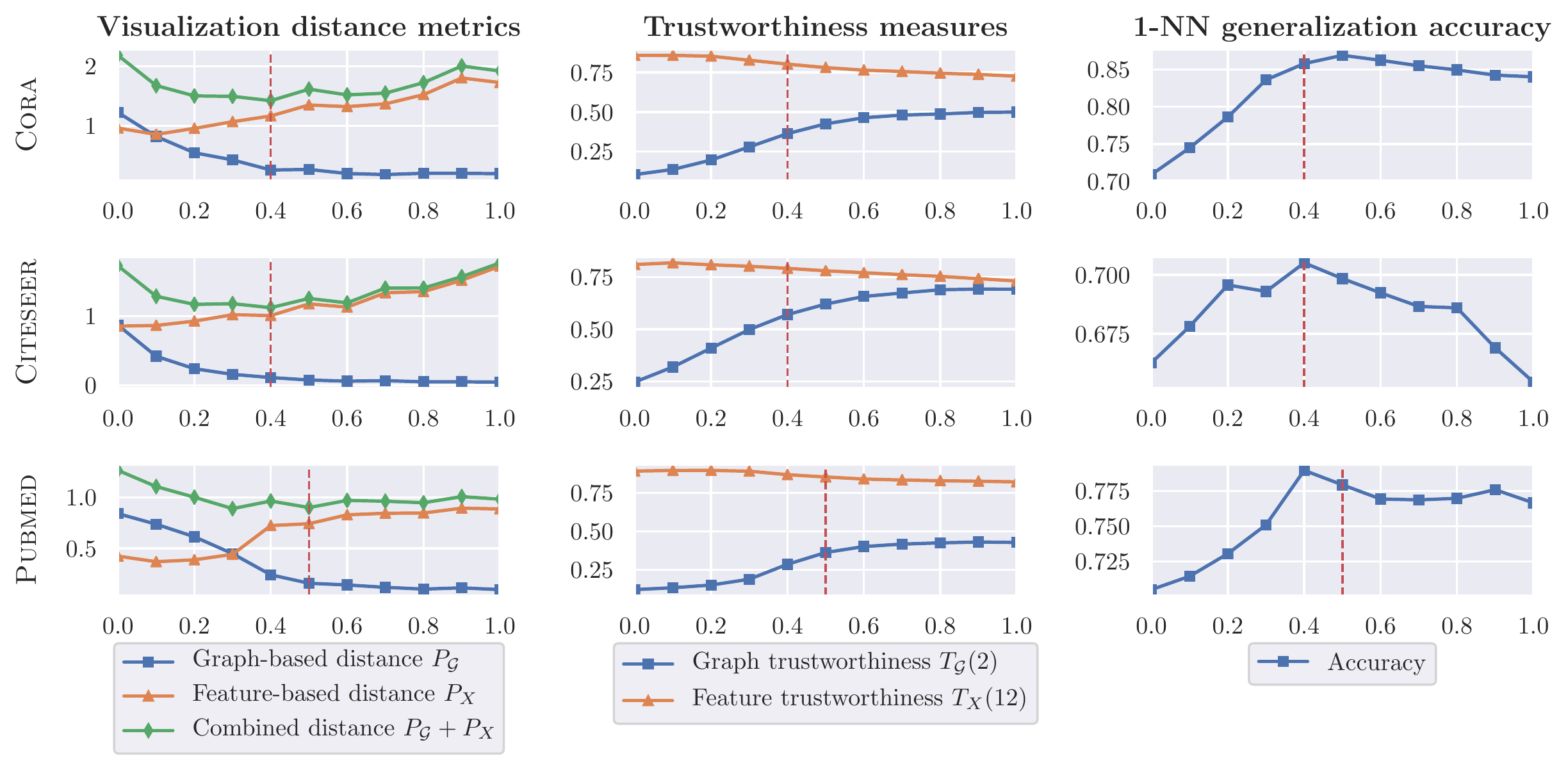}
\end{center}
\caption{Performance evaluation of different benchmarks on varying the weight of the graph clustering loss $\alpha \in [0,1]$, denoted by the $x$ axis. The red vertical line denotes $\alpha ^*$, the value of $\alpha$ that minimizes the combined distance metric $(P_\gG + P_X)$ on each dataset.}
\label{all_alphas}
\end{figure}

\textbf{Quantitative assessment.} In Figure~\ref{all_alphas}, we vary the weight of the graph clustering loss $\alpha \in [0,1]$ and report the performance of the resulting visualizations. As $\alpha$ varies, GraphTSNE tradeoffs between graph and feature trustworthiness. Based only on the trustworthiness measures, it is hard to determine the optimal value of $\alpha$, which we denote as $\alpha ^*$, since the measures have incomparable scales. Instead, we suggest setting $\alpha ^*$ to the value of $\alpha$ with the best performance on the combined distance metric $(P_\gG + P_X)$ or 1-NN generalization accuracy (if applicable). Since we do not assume the presence of class labels during training, we use the former strategy to select $\alpha ^*$.

By incorporating both graph connectivity and node features, the visualizations produced with intermediate values of $\alpha$ achieve better separation between classes and therefore, higher 1-NN generalization accuracy. This phenomenon has been well-studied in the context of semi-supervised classification (\citealp{Yang16, Kipf16, Velikovi18}) and unsupervised representation learning (\citealp{Hamilton17, Velikovi18b}). 

\begin{figure}[h]
\begin{center}
\includegraphics[width=1.0\linewidth]{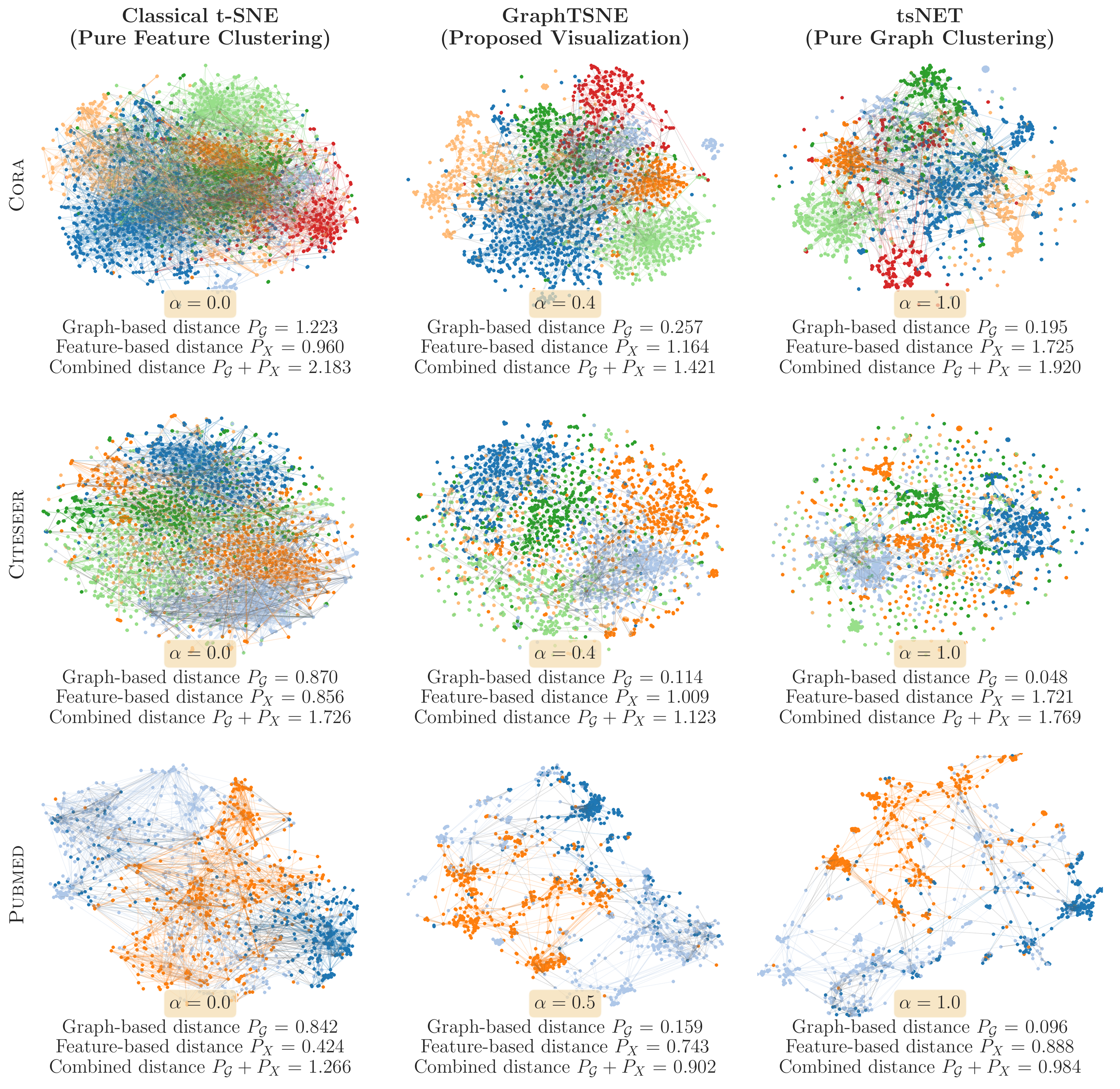}
\end{center}
\caption{Comparison of visualization techniques on benchmark datasets. Colors denote document class which are not provided during training. GraphTSNE visualizations are produced with $\alpha=\alpha ^*$.}
\label{all_plots}
\end{figure}

\textbf{Qualitative assessment.} In Figure~\ref{all_plots}, we provide a visual comparison of the citation networks under three settings of $\alpha$. At $\alpha=0$, our method reverts to pure feature clustering as in classical t-SNE. This results in long edges that crowd the layout and poorly reflect the overall graph structure. At $\alpha=1$, our method performs pure graph clustering, similar to tsNET \citep{Kruiger17}. This creates many tight clusters representing graph cliques, while arbitrarily placing disconnected nodes in the low-dimensional map. At the proposed value of $\alpha ^*$, GraphTSNE visualizations are able to accurately reflect the overall graph structure  and achieve better separation between classes.

\textbf{Training time.} Similar to t-SNE, GraphTSNE runs in $\bigO(N^2)$ time-complexity due to the computation of the input distance matrices $\mD^X$ and $\mD^\gG$. To lower the cost of this preprocessing step, we use the neighbor subsampling (NS) approach of \citet{Hamilton17} to train our GCNs using stochastic mini-batch gradient descent. After selecting a random mini-batch of nodes $\gV_B$, NS iteratively expands the mini-batch by randomly choosing $\vd(l)$ neighbors for each node at layer $l$. For our two-layer GCNs, we set $\vd=[10,15]$, yielding a receptive field size of 150 per node. After the preprocessing step, a potential speed-up, which is not in our current implementation, would be to compute the gradients in $\bigO(N\log N)$ time-complexity using tree-based approximation algorithms \citep{Maaten14}. In future work, we will explore GraphTSNE as an \textit{inductive} visualization technique that will enable its use in larger and time-evolving graph datasets.

\subsubsection*{Acknowledgments}
Xavier Bresson is supported by NRF Fellowship NRFF2017-10.

\clearpage

\bibliographystyle{graphtsne_iclr19}
\bibliography{graphtsne_iclr19}

%%%%%%%%%%%%%%%%%%%%
\appendix
%%%%%%%%%%%%%%%%%%%%
\section{Training Hyperparameters}
 \label{appendix_a}
Our method uses a residual gated graph convolutional network (GCN) \citep{Bresson18} with two graph convolutional layers. We consider different sets of hyperparameters for small datasets ($|V| \leq 10000$) and larger datasets ($|V| > 10000$). For small datasets, we train our nets with 128 hidden units per layer for 360 epochs using full-batch gradient descent. For larger datasets, we train our nets with 256 hidden units per layer for 5 epochs using mini-batch gradient descent. At each epoch, we randomly partition the dataset into 1000 batches which are expanded with neighbor subsampling. We initialize network weights with Xavier initialisation \citep{Glorot10} and use batch normalization \citep{Ioffe15}. We train using Adam \citep{Kingma15} with a learning rate of 0.00075. The learning rate scheduler has a decay factor of 1.25. Finally, we set the perplexity associated with the t-SNE loss to a default value of 30 for all experiments.

\section{Details of Trustworthiness Measures}
\label{appendix_b}
Recall the following local neighborhood definitions from Section \ref{trust}:
\begin{align*}
\eqone \\
\eqtwo \\
\eqthree
\end{align*}

The feature trustworthiness \citep{Venna06} expresses the extent to which feature neighborhoods $\sS_X$ are retained in the embedding neighborhoods $\sS_Y$. The measure is defined as:
\begin{equation} 
T_X(k) = 1 - \frac{2}{Nk(2N-3k-1)}\sum_{i=1}^N\sum_{j\in \sU(v_i,k)}(r(i,j)-k) \text{ ,}
\end{equation}
where $r(i,j)$ denotes the rank of the low-dimensional point $\vy_j$ according to the pairwise distances from a given reference point $\vy_i$. $\sU(v_i,k)$ denotes the set of points in the embedding neighborhood of $v_i$ but not in its feature neighborhood, i.e. $\sU(v_i,k) = \sS_Y(v_i,k) \backslash \sS_X(v_i,k)$.

The graph trustworthiness, adapted from \citet{Martins15}, computes the Jaccard similarity between graph neighborhoods $\sS_\gG$ and embedding neighborhoods $\sS_Y$. Given a node $v_i$ and a fixed value of $r$, the graph neighborhood is defined by the $r$-hop neighborhood of node $v_i$. For each node, set $k=|S_G(v_i,r)|$. The graph neighborhood measure is defined as:
\begin{equation} 
T_\gG(r) = \frac{1}{|\gV|}\sum_{i=1}^N \frac{|\sS_\gG(v_i,r) \cap \sS_Y(v_i,k)|}{|\sS_\gG(v_i,r) \cup \sS_Y(v_i,k)|} \text{ .}
\end{equation}

\end{document}